\documentclass[agile, final]{copernicus-agile}

\usepackage{fontawesome5} 
\usepackage{enumitem}
\hypersetup{
    colorlinks=true,
    filecolor=magenta,
    urlcolor=blue,
    linkcolor=blue,
    citecolor=black,
    }
\usepackage{tcolorbox}
\usepackage{cleveref}
\usepackage{booktabs}
\usepackage{graphicx}
\usepackage{subcaption} 
\usepackage{caption}    
\usepackage{float}      
\newcommand{\eg}{\textit{e.g.,}}

\begin{document}

\title{Leveraging LLMs and attention-mechanism for automatic annotation of historical maps}


\Author[]{Yunshuang}{Yuan}
\Author[]{Monika}{Sester}

\affil[]{Institute of Cartography and Geoinformatics, Leibniz University Hannover, Germany}





\firstpage{1}

\maketitle

\begin{abstract}
Historical maps are essential resources that provide insights into the geographical landscapes of the past. They serve as valuable tools for researchers across disciplines such as history, geography, and urban studies, facilitating the reconstruction of historical environments and the analysis of spatial transformations over time. However, when constrained to analogue or scanned formats, their interpretation is limited to humans and therefore not scalable. Recent advancements in machine learning, particularly in computer vision and large language models (LLMs), have opened new avenues for automating the recognition and classification of features and objects in historical maps. In this paper, we propose a novel distillation method that leverages LLMs and attention mechanisms for the automatic annotation of historical maps. LLMs are employed to generate coarse classification labels for low-resolution historical image patches, while attention mechanisms are utilized to refine these labels to higher resolutions.
Experimental results demonstrate that the refined labels achieve a high recall of more than 90\%. Additionally, the intersection over union (IoU) scores—84.2\% for Wood and 72.0\% for Settlement—along with precision scores of 87.1\% and 79.5\%, respectively, indicate that most labels are well-aligned with ground-truth annotations. Notably, these results were achieved without the use of fine-grained manual labels during training, underscoring the potential of our approach for efficient and scalable historical map analysis.

\textbf{Submission Type.} Model, Algorithm


\textbf{BoK Concepts.} Image Classification, Semantic Enrichment, Image Segmentation  


\end{abstract}


\section{Introduction}

Historical maps provide invaluable insights into landscape development and land use changes over time, as their information reaches far back into the past. However, their analogue or scanned formats limit accessibility and usability for modern applications. To make them accessible in a scalable and automatic fashion, the contents of these maps have to be described explicitly - which is often provided in terms of general metatdata (\eg~general information about semantic contents) or also more precisely in terms annotations, \eg~ of hierarchical structure or geometric relationships of the objects in the maps.
The availability of such additional data has many benefits: on the one hand, historical map data can be queried and inspected in a more convenient way (\eg~via keywords, text or parameters); on the other hand, the description can also serve to explain the content of the map to visually impaired or blind people - and thus allow a broader accessibility of the data (\cite{robinson2024using}).

To enhance the accessibility of historical maps, traditional methods rely on digitization of the content in geographic information systems (GIS) and manually extract and store geographic features, along with their structures and relationships~\citep{salt_marsg,Levin_Kark_Galilee_2010,san-antonio-gomez_urban_2014,picuno_investigating_2019,tonolla_seven_2021}. While effective, this process is highly time-consuming and lacks scalability. To improve efficiency, some studies~\citep{leyk_segmentation_2010,Uhl_Leyk_2021,uhl_automated_2020} have employed computer vision techniques that statistically analyze pixel contexts within maps based on prior knowledge to assign semantic labels to pixels or image patches. Although these methods automate the annotation process, their scalability remains limited, as statistical models often fail to generalize across maps with varying visual characteristics.
In contrast, deep learning-based semantic segmentation~\citep{Csurka2023SemanticIS,Yuan_gevbev2023,heitzler_cartographic_2020,ekim_automatic_2021,wu_closer_2022,wu_leveraging_2022}, which assigns semantic labels at the pixel level, offers a more scalable solution. However, these approaches require extensive training data and ground-truth annotations, which are particularly costly and labor-intensive to generate for historical maps. To address this challenge, prior research has explored methods to reduce manual annotation efforts, such as domain adaptation techniques~\citep{wu_domain_2023} and weak supervision through age-tracing starting from a single annotated map sheet~\citep{yuan2025hismap}. While these strategies improve semantic segmentation performance with limited ground-truth data, they still rely on tedious manual annotations. Moreover, these models often struggle to generalize across maps with significant domain differences, such as those produced by different cartographers or depicting distinct geographical regions.

Recent advances in AI, particularly in Deep Learning (DL) and LLMs, have enabled powerful new possibilities for data interpretation. The general knowledge embedded in LLMs can be distilled into smaller DL models without requiring manual annotations. More importantly, this approach can be efficiently implemented at scale. In this paper, we leverage LLMs and the attention mechanism~\citep{NIPS2017_attn} to develop a \textit{knowledge-distillation framework for generating semantic annotations for historical maps} automatically. Specifically, we utilize LLMs to generate semantic labels for large cropped images (\eg~$384\times384$ pixels) and then train an attention-based image classification model using these labels. By employing the attention mechanism, the model identifies clues associated with specific classes within the images. These clues, represented as attention weight maps, can be further used as annotations for more fine-grained image patches (\eg~$64\times64$ pixels). This approach is expected to make image patch-level annotation more cost-efficient and scalable across various map styles.

These annotations have a wide range of applications. For instance, they can be directly utilized to characterize and describe the content of historical maps. Descriptions may involve simply enumerating the object classes present on the map (\eg~land use categories such as settlements or forests), while more detailed analyses could include specifying the sizes and extents of these objects. Furthermore, the generated patch-wise semantic labels can serve as supervision for training semantic segmentation models, enabling the refinement of labels from patch-level to pixel-level accuracy.
Building upon these multi-resolution semantic labels, it becomes possible to extract additional information such as geometric delineations of objects, verbal descriptors (\eg~"elongated village"), and spatial relationships (\eg~"a settlement beside a river"). Moreover, these enriched annotations can facilitate the construction of scene graphs~\citep{janowicz2022know}, representing the spatial and semantic relationships between objects within the map. In this paper, we will provide a proof-of-concept of the first part of this knowledge-distillation chain, namely the automatic image patch-level annotation.

\section{Method}

\subsection{LLM-based label generation}
\begin{figure}[t]
    \includegraphics[width=0.8\linewidth]{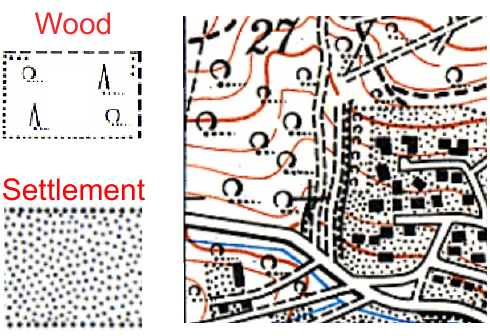}
    \caption{Prompt image for LLM.}
    \label{fig:prompt_image}
\end{figure}

We demonstrate the distillation concept using two classification categories: Wood and Settlement. To achieve this, we prompt the LLM to generate image-level class labels. For each image, a prompt image is created, as illustrated in \cref{fig:prompt_image}, alongside a corresponding prompt text, shown in \cref{fig:llm_prompting}. The LLM then assigns classification labels to each cropped image patch of the historical maps. In the example, the LLM correctly identifies the image patch as containing both Wood and Settlement classes.

\begin{figure}
\centering
  \begin{minipage}[t]{0.5\textwidth} 
    \begin{tcolorbox}[colback=gray!5!white, colframe=gray!75!black, title=Prompt]
    \textit{On the left side is some examples of the symbols of a historical map including the class 
       Wood and Settlement. On the right side is an image of historical map patch. 
       For the right image, please answer the following question with Yes or No and give reasons for the answer: \\
        1. Does the image contain Wood? \\
        2. Does the image contain Settlement? \\        
        Formatting the answer with the following structure:\\
        1. **Wood?** [Yes/No] : [reason]\\
        2. **Settlement?** [Yes/No] : [reason]  
    }
    \end{tcolorbox}
  \end{minipage}%
  \hspace{0.01\textwidth} 
  \begin{minipage}[t]{0.5\textwidth} 
    \begin{tcolorbox}[colback=gray!5!white, colframe=gray!75!black, title=Answer from ChatGPT-4o]
    \textit{
        1. **Wood?** Yes: In the right image, clusters of closely spaced, small circular symbols (as seen in the wood example) can be observed. These represent the wood class.\\
        2. **Settlement?** Yes: The map contains dotted patterns that represent settlements."  
    }
    \end{tcolorbox}
  \end{minipage}
\caption{LLM prompting on an example of a historical map.}
\label{fig:llm_prompting}
\end{figure}

\subsection{Framework of image classification}

\begin{figure*}
    \centering
    \includegraphics[width=\linewidth]{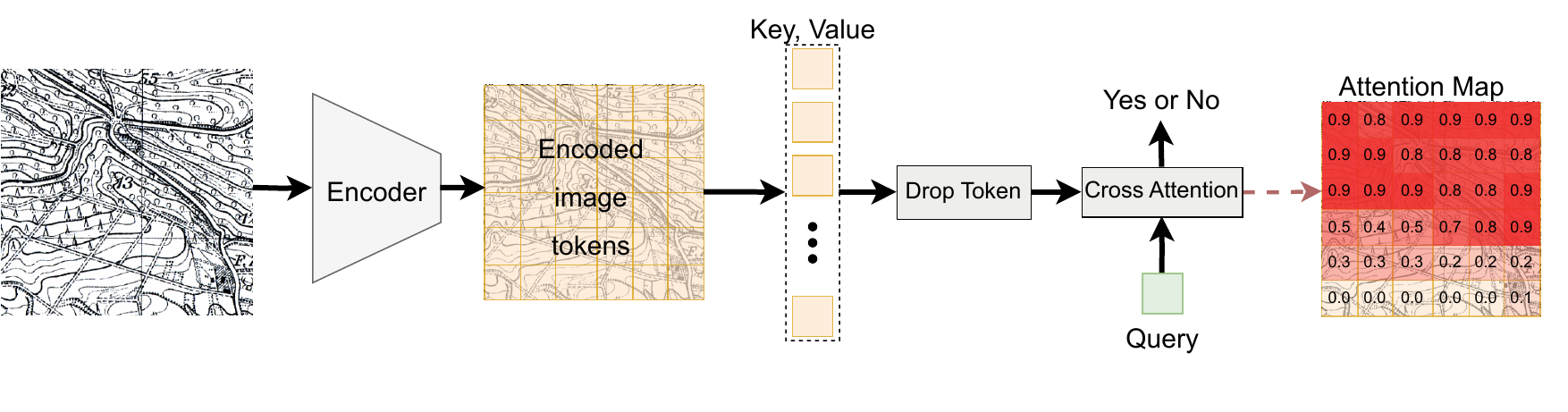}
    \caption{Framework for attention-based image classification. The input image features are first extracted by the \textit{Encoder} into image tokens, with a subset discarded by the \textit{Drop Token} module. The remaining tokens are processed by the \textit{Cross-Attention} module to produce the final binary classification result. Post-training, the learned attention weights from the \textit{Cross-Attention} module are used to generate the \textit{Attention Map}.}
    \label{fig:pipeline}
\end{figure*}

\begin{figure*}
    \centering
    \includegraphics[width=\linewidth]{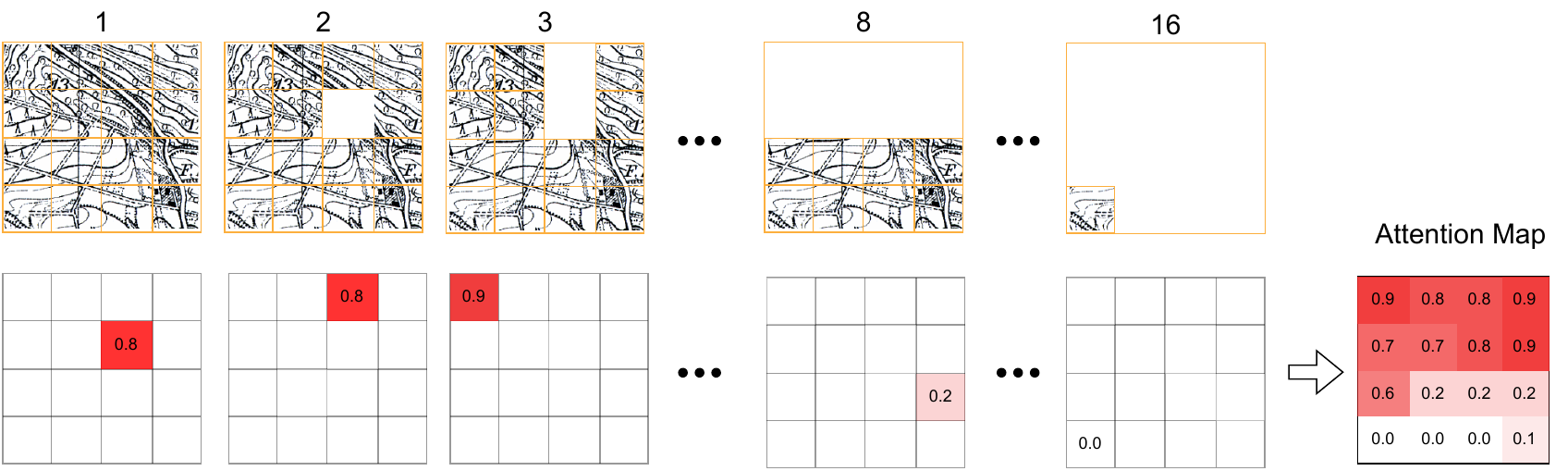}
    \caption{An example of attention map generation with 16 image tokens. The final attention map is generated with 16 forward runs of the trained model. Each column indicates one round. The white squares in the first raw indicate that the corresponding token is dropped (features are set to zeros). The red squares in the second row show the selected maximum attention weight in each forward round, eventually composing the Attention Map.}
    \label{fig:attn_map}
\end{figure*}

\begin{figure*}[t]
    \centering
    \begin{subfigure}{0.49\textwidth} 
        \centering
        \includegraphics[width=\linewidth]{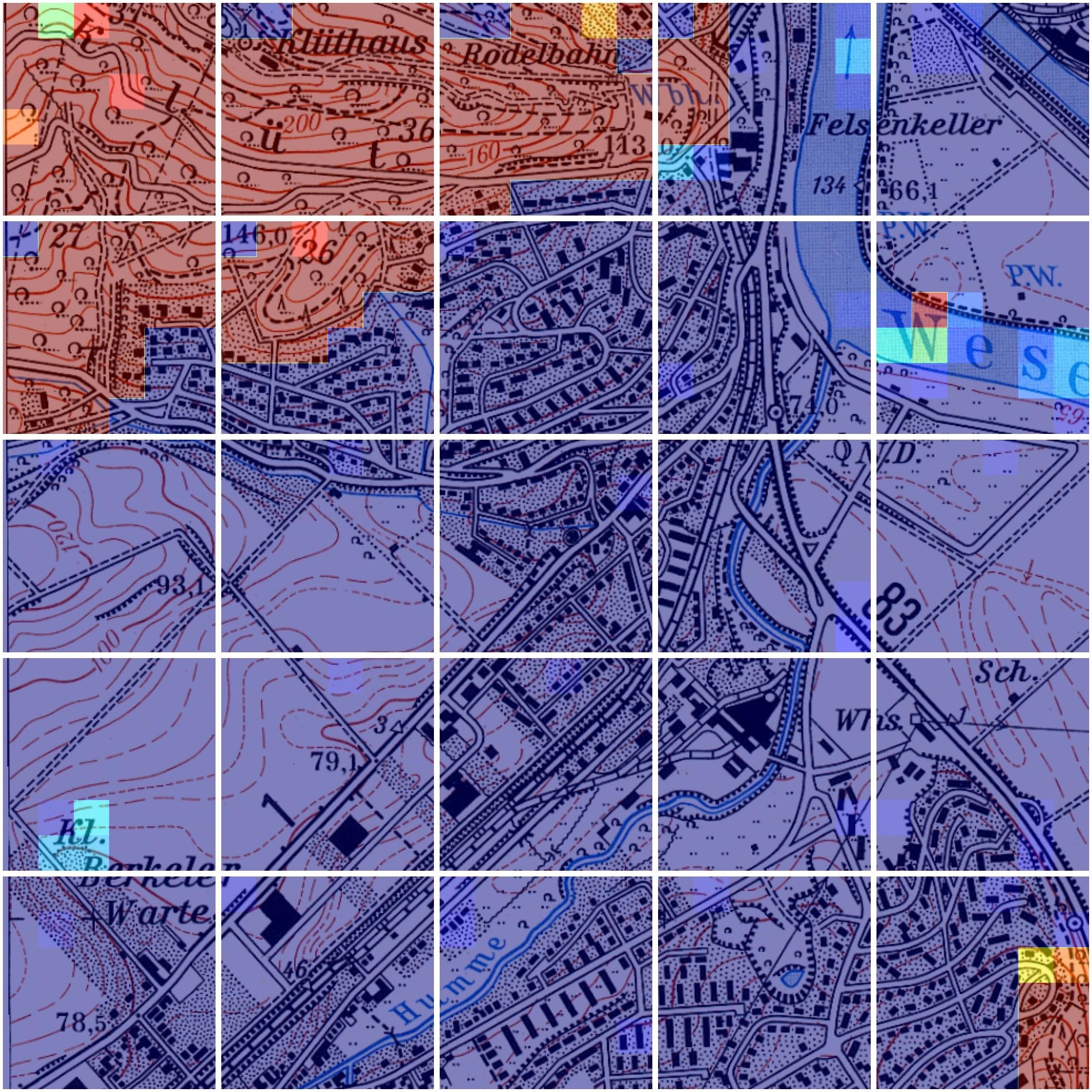} 
        \caption{Wood}
        \label{fig:wood}
    \end{subfigure}
    \hfill
    \begin{subfigure}{0.49\textwidth} 
        \centering
        \includegraphics[width=\linewidth]{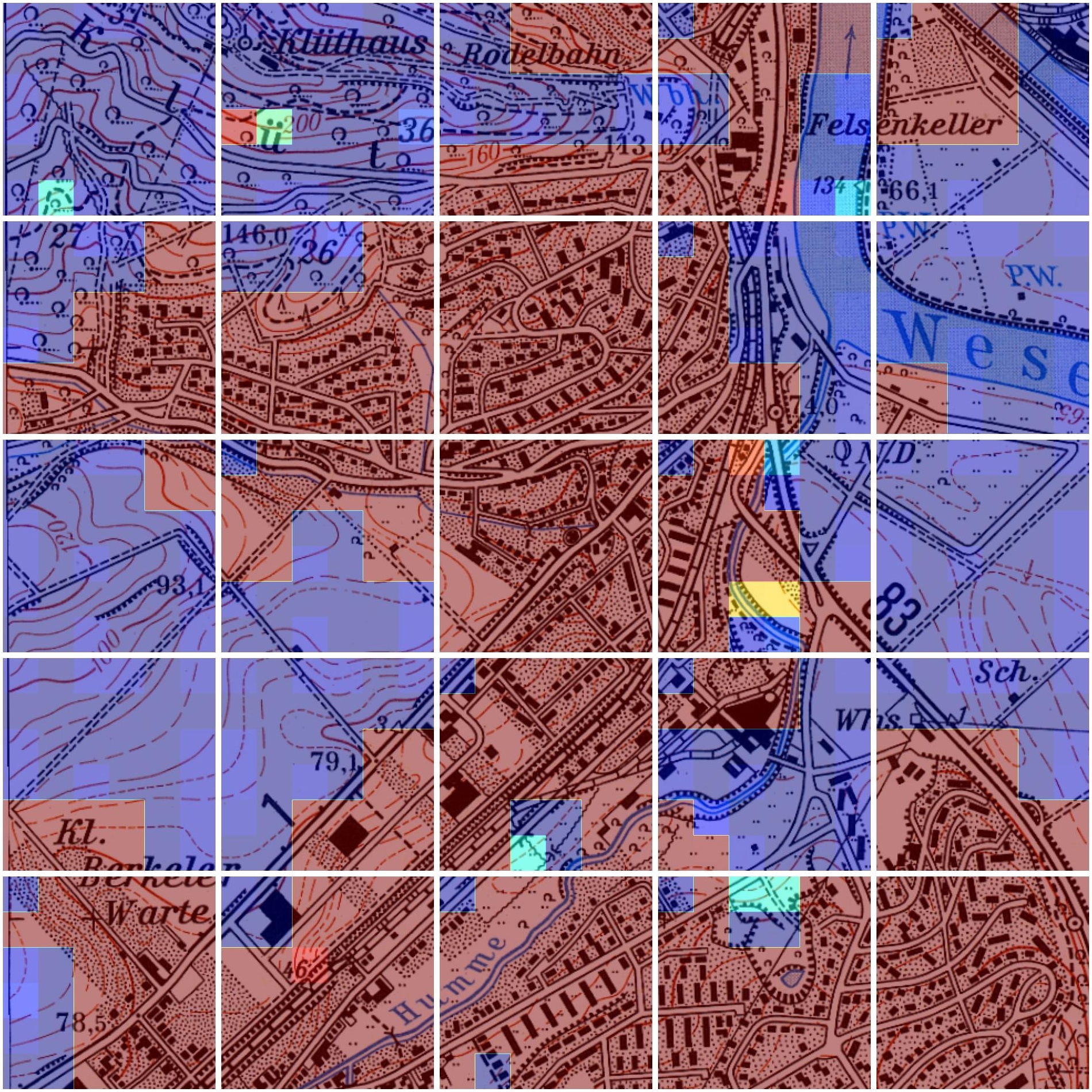} 
        \caption{Settlement}
        \label{fig:settlement}
    \end{subfigure}
    \begin{subfigure}{\textwidth} 
    \centering
    \includegraphics[width=0.48\linewidth]{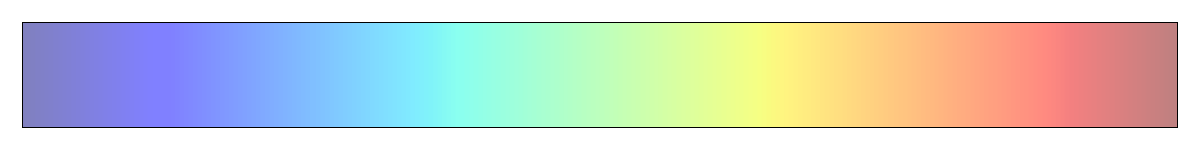} 
    \caption{Attention weight scale. Blue to red: low attention weight (0.0) to high attention weight (1.0).}
    \label{fig:scale}
    \end{subfigure}
    \caption{Example of attention maps overlay on a $5\times 5$ grid of input images. Each image is covered by a $6\time 6$ grid of attention weights.}
    \label{fig:res_attn_maps}
\end{figure*}

The image classification framework, illustrated in \cref{fig:pipeline}, comprises three main modules: \textit{Encoder}, \textit{Drop Token}, and \textit{Cross Attention}~\citep{chen2021crossvit}. The \textit{Encoder} first encodes the input image into a set of image tokens, each represents the extracted features of a specific image patch. In the \textit{Drop Token} module, a subset of these tokens is randomly discarded based on a predefined probability $0<p<1$. This random dropping mechanism prevents the attention module from focusing exclusively on easily classifiable patches, thereby promoting more comprehensive feature learning.
The remaining tokens are subsequently passed to the \textit{Cross Attention} module to produce the final binary classification results. After model training, the attention weights from the \textit{Cross Attention} module are utilized to generate an attention map, which can support further analysis of historical maps.

\textbf{Encoder}
The \textit{Encoder} consists of six blocks, each comprising two convolutional layers followed by a max-pooling layer that reduces the spatial dimensions by half. Each convolutional layer employs the \textit{ReLU} activation function to introduce non-linearity.
Given the input image of dimensions $H\times W \times 3$, the encoder extracts feature maps with dimensions $\frac{H}{64}\times \frac{W}{64} \times C$, where $C$ denotes the number of output feature channels. This transformation results in feature maps of dimensions $M\times N\times C$, corresponding to $L=M\cdot N$ encoded image tokens.

\textbf{Drop Token}
For each of the $L$ input tokens, we apply a Bernoulli distribution with a probability of $p$ to randomly drop a subset of tokens, encouraging the model to learn from more challenging tokens and promoting robust feature learning. However, the randomness introduced by this process results in a variable number of remaining tokens $S$, which can affect the stability of the subsequent attention module. Since this module aggregates attention weights across all input tokens to generate the final classification results, fluctuations in token count may lead to inconsistencies in model performance. To address this issue, we normalize the remaining token features by scaling them with the inverse of the retention probability, $\frac{1}{1-p}$, before passing them to the attention module. This normalization ensures that the expected contribution of each token remains consistent, thereby stabilizing the attention mechanism and improving classification reliability.

\textbf{Cross Attention}
Given the remaining tokens as key and value $K,V\in \mathbb{R}^{S\times C}$, a randomly initialized query $Q\in \mathbb{R}^{1\times C}$, and positional embeddings $P_q\in \mathbb{R}^{1\times C}$ and $P_{kv}\in \mathbb{R}^{S\times C}$ for $Q$ and $K,V$, respectively, the cross attention is formulated as follows:

\begin{align}
\centering
    Q &= linear_q(Q + P_q) \\
    K &= linear_k(K + P_{kv}) \\
    V &= linear_v(V + P_{kv}) \\
    W &= softmax(\frac{Q \cdot K^T}{\sqrt{C}}) \\
    Q &= linear(W \cdot V ) \label{eq:out_q}
\end{align}

Here, $linear$ denotes fully connected layer, and $softmax$ normalizes the attention weights in $W$ to the range of $[0, 1]$. The attention weight $W$ has a shape of $1\times S$, while the output query features in \cref{eq:out_q} are of shape $Q\in \mathbb{R}^{1\times C}$. Finally, the updated $Q$ is passed through a classification head, consisting of a single linear layer, to produce the final classification result.

\subsection{Attention map as annotation}

The attention weights produced by the \textit{Cross Attention} module often focus on a small subset of tokens when making the final classification. For instance, as shown in \cref{fig:attn_map}, even if approximately half of the input image contains the Wood class, the attention module may only need to attend to one or a few representative Wood patches to correctly determine whether the entire image contains the Wood class. To generate an attention map that assigns high weights and highlights the entire foreground area, we run the model for $L$ iterations, selecting the maximum attention weight from each iteration to update the final attention map.
Mathematically, given a set of image tokens  $I=\{I_i | i\in \{1, \dots, L\}\}$ and an initially empty attention map $A=\emptyset$, this process can be formulated as follows:
\begin{enumerate}
    \item Generate attention weights: Use the trained model to compute attention weights $W=\{W_i | i\in \{1, \dots, L\}\}$.
    \item Select Maximum Attention Weight: Identify the highest attention weight $W_i = max(W)$, and its corresponding index $j=argmax(W)$. Update the attention map as $A=A\cup {W_i}$.
    \item Drop the Selected Token: Remove the token $I_j$ for the next iteration.
\end{enumerate}
These steps are repeated until all image tokens have been processed. An illustration of this iterative process is shown in \cref{fig:attn_map}. For example, in the first iteration (column 1), all tokens are passed through the attention module, yielding a maximum attention weight of 0.8. In the second iteration, the image token corresponding to the previous maximum weight is removed (or its features are set to zero). The new maximum attention weight 0.8 corresponds to a different location. As this process continues, we progressively build an attention map that highlights the entire relevant region, as the \textit{Attention Map} depicted in the final column of \cref{fig:attn_map}. These weight maps can then be used annotations for other tasks, such as semantic segmentation.

\subsection{Experimental settings}
\textbf{Dataset}
We conduct our experiments using map sheets published by the Lower Saxony Mapping Agency (LGLN). To reduce training time while effectively demonstrating our knowledge distillation approach from large language models (LLMs), we selected maps from the years 1973 to 1975. Three map sheets (3821, 3822, and 3921) covering the Hameln area were used for training, while one map sheet (3922) was reserved for evaluation.
Each map sheet was cropped into images of size $384 \times 384$ pixels. Using an LLM, we annotated each cropped image with a binary label for each foreground class. Given that LLM-generated annotations may not be fully accurate, we visualized the labels in an interactive interface, allowing human annotators to efficiently correct errors by simply clicking to flip incorrect labels. As the majority of labels were accurate, this correction process was highly efficient, typically requiring less than one minute per map sheet.

\textbf{Training}
The cropped images result in encoded image tokens of shape $6\time 6$, corresponding to $36$ tokens per image. In the \textit{Drop Token} module, we empirically set the drop probability to $p=0.2$ and the number of encoding channels to $C=512$.
The model is trained with Focal loss for 100 epochs. Optimization was performed with the Adam optimizer, using a learning rate of $5\cdot 10^{-4}$ and a warm-up period of 5 epochs. We trained two separate models to identify the foreground classes: Wood and Settlement.

\section{Result and evaluation}
\subsection{Qualitative result}
The generated attention maps for some image patches are presented in \cref{fig:res_attn_maps}, while results for the whole map sheets are in the Appendix.  
The attention weights are visualized using a color gradient (\cref{fig:scale}), where red and blue represent high and low attention weights, respectively. Two separate models are trained to produce \cref{fig:wood} and \cref{fig:settlement}, one for each class considered as foreground class. It can be observed that the majority of image patches are correctly classified, with foreground classes highlighted by high attention weights.

Despite these promising results, the classification is not flawless, and two primary issues are evident. First, the model tends to misclassify certain background patches as foreground. For instance, in \cref{fig:wood}, the patch at row 2 and column 5, denoted as $P(2, 5)$, is incorrectly classified, while in \cref{fig:settlement}, $P(1, 2)$ show similar errors. This issue is particularly noticeable at the image borders, such as patches $P(1, 1), P(1, 2), P(1, 3), P(2, 1), P(2, 2)$ in \cref{fig:wood}.

Second, the model struggles to accurately delineate the boundaries between foreground and background classes. It often extends the foreground classification beyond the actual boundaries defined by the ground truth. For example, in \cref{fig:wood}, the image patch $P(2, 1)$ incorrectly includes parts of a settlement area within the wood class. Similar boundary misclassifications are also visible in \cref{fig:settlement}.

We hypothesize that these issues arise from the large receptive fields of the convolutional kernels, which expand with increasing convolutional layers. While additional layers are necessary for improved feature embedding and capturing broader contextual information, they also increase the likelihood of neighboring patches—those close to the foreground but belonging to the background—sharing similar features with the foreground. Consequently, these patches are more prone to being misclassified as foreground.

\begin{table}[t]
    \centering
    \caption{Down-sampled: patch-wise ($64\times 64$ pixels) classification results at attention weight threshold of 0.5.}
    \label{tab:quantative_down}
    \begin{tabular}{l |c |c| c}
    \toprule
             & IoU & Precision & Recall\\\hline
        Wood &  0.824 & 0.871 &  0.939\\\hline
        Settlement & 0.720 & 0.795 & 0.880\\
        \bottomrule
    \end{tabular}

\end{table}

\begin{table}[t]
    \centering
    \caption{Up-sampled: pixel-wise classification results at attention weight threshold of 0.5.}
    \label{tab:quantative_up}
    \begin{tabular}{l |c |c| c}
    \toprule
             & IoU & Precision & Recall\\\hline
        Wood &  0.781 & 0.796 &  0.975\\\hline
        Settlement & 0.474 & 0.482 & 0.968\\
        \bottomrule
    \end{tabular}
    
\end{table}

\begin{figure}[t]
    \centering
    \includegraphics[width=\linewidth]{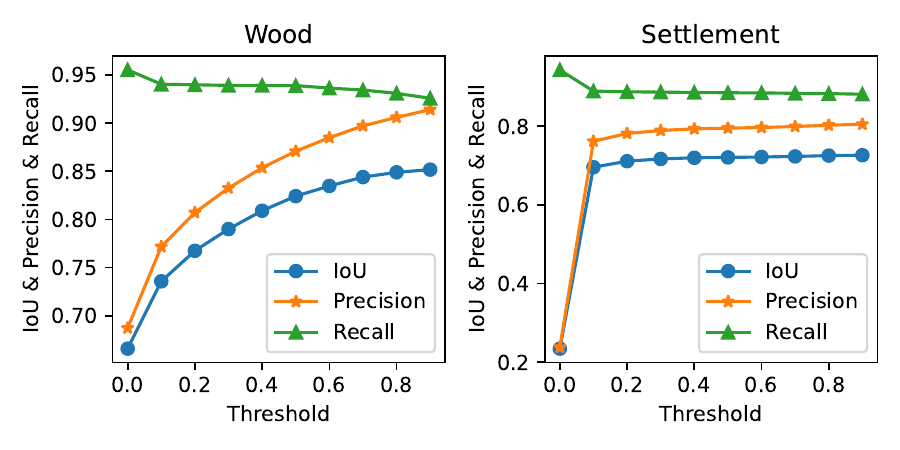}
    \caption{Down-sampled: patch-wise ($64\times 64$ pixels) classification results with different thresholds for attention weights.}
    \label{fig:thrs_down}
\end{figure}

\begin{figure}[t]
    \centering
    \includegraphics[width=\linewidth]{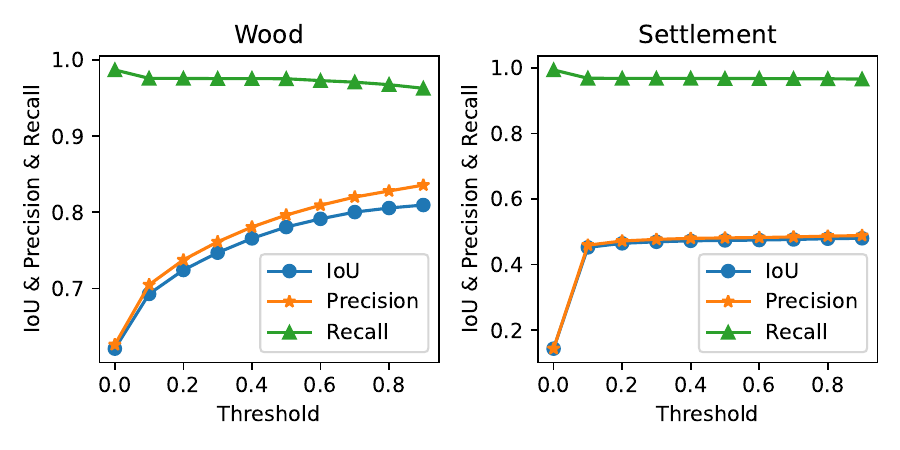}
    \caption{Up-sampled: classification results with different thresholds for attention weights.}
    \label{fig:thrs_up}
\end{figure}

\subsection{Quantitative result}
To quantitatively evaluate the proposed framework, we compare the generated attention maps, denoted as $A$, with the ground-truth semantic segmentation labels, denoted as $Y$. However, as the attention maps and the semantic labels have different resolutions, an alignment was necessary for a fair comparison. Each attention weight $W_i$ in the attention map corresponds to an image patch of size $64\times 64$ pixels, whereas the ground-truth $Y$ provides pixel-wise labels. To harmonize the resolutions, we employed two approaches: down-sampling the pixel-wise labels $Y$ or up-sampling the patch-wise attention maps $A$. 

For the \textit{down-sampled comparison}, $Y$ is divided into $64\times 64$ tiles. Each tile is assigned as foreground if it contains any foreground pixels; otherwise, it is labeled as background. For the \textit{up-sampled comparison},
each attention weight $W_i$ is treated as a classification confidence score and uniformly assigned to all pixels within their corresponding $64\times 64$ image patch. Repeating this process across the attention map results in an up-sampled attention map matching the resolution of $Y$.

In both comparisons, we apply a threshold $0<\sigma <1$ to produce the final classification results. Specifically, pixels in attention maps with values greater than $\sigma$ are classified as foreground, while those with values less than or equal to $\sigma$ are classified as background. Using these classification results alongside the ground-truth labels of matching resolution, we compute the Intersection over Union (IoU), precision, and recall as evaluation metrics. 

The down-sampled comparison results, presented in \cref{tab:quantative_down}, indicate that most image patches ($64\times 64$) are correctly classified. For example, the precision reaches $87.1\%$ for wood and $79.5\%$ for settlement class. The recall values are even higher, indicating the most foreground patches are successfully identified. The wood class outperforms settlement, likely due to its higher spatial coverage, which increases the likelihood of patches being classified as foreground.

In contrast, the up-sampled comparison (\cref{tab:quantative_up}) yielded lower IoU and precision. This decline can be attributed to the coarse-grained predictions failing to align precisely with the fine-grained ground-truth labels, even after up-sampling. Furthermore, recall remained significantly higher compared to IoU and precision. This trend aligns with qualitative observations where background patches near foreground boundaries were often misclassified as foreground. The elevated recall suggests that the model reliably identifies most instances of the target foreground class, which is particularly beneficial for applications such as content-based historical map retrieval.

We also evaluated the model's performance across various attention weight thresholds, as shown in \cref{fig:thrs_down} and \cref{fig:thrs_up}. The results demonstrate that increasing the threshold improves IoU and precision, though it slightly reduces recall. This effect was less pronounced for the settlement class, where metrics stabilized at lower thresholds but achieved lower maximum values—IoU of $0.72$ (down-sampled) and $0.47$ (up-sampled), and precision of $0.80$ (down-sampled) and $0.48$ (up-sampled).
In contrast, the wood class demonstrated superior performance, with IoU reaching $0.85$ and $0.81$ for down- and up-sampled comparisons, respectively. Precision peaked at $0.93$ (down-sampled) and $0.84$ (up-sampled) when the threshold was set to $\sigma=0.9$.

\section{Conclusion and future work}
\vspace{-0.4cm}
In this paper, we proposed a method for distilling the knowledge of LLMs into compact, attention-based models for recognizing content in historical maps. Specifically, we utilized LLMs to annotate large historical image patches by determining the presence of specific semantic classes. These labelled patches were then employed to train an attention-based classification model.
By leveraging the attention weights from the trained model, we were able to trace the location of target semantic classes within the image, enabling the refinement of semantic labels at higher resolutions. Experimental results demonstrate that these refined semantic labels closely align with ground-truth pixel-wise annotations, achieving a high recall rate of more than $90\%$. Nonetheless, some misclassifications were observed at class boundaries, where patches near edges were incorrectly labelled as foreground. Future work will investigate the influence of such label noise on subsequent processes, as described in the introduction. 
In the current approach, the LLM-generated labels had an accuracy of approx. 70\%. While a quick correction of those labels is possible due to the large patch-sizes, future work will also try to improve the labelling results.  

Also, we will research possibilities to improve the current results. In our experiments, we adopted a fixed input patch size of $384\times 384$ pixels and an output patch size of $64\times 64$ pixels. Future work could explore the effects of varying input and output patch sizes to further optimize model performance. Additionally, hierarchical model architectures could be investigated to progressively reduce output patch sizes, enabling more fine-grained semantic labelling. While this study focused on binary classification—requiring separate models for each semantic class—future research could aim to develop unified models capable of multi-class classification within a single framework. In this work, we demonstrated the feasibility of generating fine-grained semantic labels from coarse labels using a simple encoder. As a next step, future research could explore leveraging deeper, pre-trained foundation models—such as Vision Transformers~\citep{vit}, and CLIP~\citep{clip}—to enhance the performance.

\subsection*{Acknowledgements}

The research in this project is conducted in the context of the Gauss-Project, funded by the Federal Agency of Cartography and Geodesy (BKG), Frankfurt, Germany.

Generative AI were used to assist with language refinement. The authors take full responsibility for the content of this manuscript.

\bibliographystyle{copernicus-agile}
\bibliography{example.bib}

\begin{thebibliography}{22}
\providecommand{\natexlab}[1]{#1}
\providecommand{\url}[1]{{\tt #1}}
\providecommand{\urlprefix}{URL }
\expandafter\ifx\csname urlstyle\endcsname\relax
  \providecommand{\doi}[1]{https://doi.org/\discretionary{}{}{}#1}\else
  \providecommand{\doi}{https://doi.org/\discretionary{}{}{}\begingroup \urlstyle{rm}\Url}\fi

\bibitem[{Bromberg and Bertness(2005)}]{salt_marsg}
Bromberg, K.~D. and Bertness, M.~D.: Reconstructing New England salt marsh losses using historical maps, Estuaries, 28, 823--832, \urlprefix\url{https://doi.org/10.1007/BF02696012}, 2005.

\bibitem[{Chen et~al.(2021)Chen, Fan, and Panda}]{chen2021crossvit}
Chen, C.-F.~R., Fan, Q., and Panda, R.: {CrossViT: Cross-Attention Multi-Scale Vision Transformer for Image Classification}, in: International Conference on Computer Vision (ICCV), 2021.

\bibitem[{Csurka et~al.(2023)Csurka, Volpi, and Chidlovskii}]{Csurka2023SemanticIS}
Csurka, G., Volpi, R., and Chidlovskii, B.: Semantic Image Segmentation: Two Decades of Research, Found. Trends Comput. Graph. Vis., 14, 1--162, \urlprefix\url{https://api.semanticscholar.org/CorpusID:253028117}, 2023.

\bibitem[{Dosovitskiy et~al.(2020)Dosovitskiy, Beyer, Kolesnikov, Weissenborn, Zhai, Unterthiner, Dehghani, Minderer, Heigold, Gelly, Uszkoreit, and Houlsby}]{vit}
Dosovitskiy, A., Beyer, L., Kolesnikov, A., Weissenborn, D., Zhai, X., Unterthiner, T., Dehghani, M., Minderer, M., Heigold, G., Gelly, S., Uszkoreit, J., and Houlsby, N.: An Image is Worth 16x16 Words: Transformers for Image Recognition at Scale, ArXiv, abs/2010.11929, \urlprefix\url{https://api.semanticscholar.org/CorpusID:225039882}, 2020.

\bibitem[{Ekim et~al.(2021)Ekim, Sertel, and Kabaday?}]{ekim_automatic_2021}
Ekim, B., Sertel, E., and Kabaday?, M.~E.: Automatic {Road} {Extraction} from {Historical} {Maps} {Using} {Deep} {Learning} {Techniques}: {A} {Regional} {Case} {Study} of {Turkey} in a {German} {World} {War} {II} {Map}, ISPRS International Journal of Geo-Information, 10, 492, \doi{10.3390/ijgi10080492}, 2021.

\bibitem[{Heitzler and Hurni(2020)}]{heitzler_cartographic_2020}
Heitzler, M. and Hurni, L.: Cartographic reconstruction of building footprints from historical maps: {A} study on the {Swiss} {Siegfried} map, Transactions in GIS, 24, 442--461, \doi{10.1111/tgis.12610}, \_eprint: https://onlinelibrary.wiley.com/doi/pdf/10.1111/tgis.12610, 2020.

\bibitem[{Janowicz et~al.(2022)Janowicz, Hitzler, Li, Rehberger, Schildhauer, Zhu, Shimizu, Fisher, Cai, Mai et~al.}]{janowicz2022know}
Janowicz, K., Hitzler, P., Li, W., Rehberger, D., Schildhauer, M., Zhu, R., Shimizu, C., Fisher, C., Cai, L., Mai, G., et~al.: Know, Know Where, KnowWhereGraph: A densely connected, cross-domain knowledge graph and geo-enrichment service stack for applications in environmental intelligence, AI Magazine, 43, 30--39, 2022.

\bibitem[{Levin et~al.(2010)Levin, Kark, and Galilee}]{Levin_Kark_Galilee_2010}
Levin, N., Kark, R., and Galilee, E.: Maps and the settlement of southern Palestine, 1799-1948: an historical/GIS analysis, Journal of Historical Geography, 36, 1--18, \doi{https://doi.org/10.1016/j.jhg.2009.04.001}, 2010.

\bibitem[{Leyk(2010)}]{leyk_segmentation_2010}
Leyk, S.: Segmentation of {Colour} {Layers} in {Historical} {Maps} {Based} on {Hierarchical} {Colour} {Sampling}, in: Graphics {Recognition}. {Achievements}, {Challenges}, and {Evolution}, edited by Ogier, J.-M., Liu, W., and Llad?s, J., pp. 231--241, Springer, Berlin, Heidelberg, \doi{10.1007/978-3-642-13728-0_21}, 2010.

\bibitem[{Picuno et~al.(2019)Picuno, Cillis, and Statuto}]{picuno_investigating_2019}
Picuno, P., Cillis, G., and Statuto, D.: Investigating the time evolution of a rural landscape: {How} historical maps may provide environmental information when processed using a {GIS}, Ecological Engineering, 139, 105\,580, \doi{10.1016/j.ecoleng.2019.08.010}, 2019.

\bibitem[{Radford et~al.(2021)Radford, Kim, Hallacy, Ramesh, Goh, Agarwal, Sastry, Askell, Mishkin, Clark, Krueger, and Sutskever}]{clip}
Radford, A., Kim, J.~W., Hallacy, C., Ramesh, A., Goh, G., Agarwal, S., Sastry, G., Askell, A., Mishkin, P., Clark, J., Krueger, G., and Sutskever, I.: Learning Transferable Visual Models From Natural Language Supervision, in: International Conference on Machine Learning, \urlprefix\url{https://api.semanticscholar.org/CorpusID:231591445}, 2021.

\bibitem[{Robinson and Griffin(2024)}]{robinson2024using}
Robinson, A.~C. and Griffin, A.~L.: Using {AI} to Generate Accessibility Descriptions for Maps, Abstracts of the ICA, 7, 139, 2024.

\bibitem[{San Antonio~Gómez et~al.(2014)San Antonio~Gómez, Velilla, and Manzano~Agugliaro}]{san-antonio-gomez_urban_2014}
San Antonio~Gómez, C., Velilla, C., and Manzano~Agugliaro, F.: Urban and landscape changes through historical maps: {The} {Real} {Sitio} of {Aranjuez} (1775-2005), a case study, Computers, Environment and Urban Systems, 44, 47--58, \doi{10.1016/j.compenvurbsys.2013.12.001}, 2014.

\bibitem[{Tonolla et~al.(2021)Tonolla, Geilhausen, and Doering}]{tonolla_seven_2021}
Tonolla, D., Geilhausen, M., and Doering, M.: Seven decades of hydrogeomorphological changes in a near-natural ({Sense} {River}) and a hydropower-regulated ({Sarine} {River}) pre-{Alpine} river floodplain in {Western} {Switzerland}, Earth Surface Processes and Landforms, 46, 252--266, \doi{10.1002/esp.5017}, \_eprint: https://onlinelibrary.wiley.com/doi/pdf/10.1002/esp.5017, 2021.

\bibitem[{Uhl et~al.(2020)Uhl, Leyk, Chiang, Duan, and Knoblock}]{uhl_automated_2020}
Uhl, J.~H., Leyk, S., Chiang, Y.-Y., Duan, W., and Knoblock, C.~A.: Automated {Extraction} of {Human} {Settlement} {Patterns} {From} {Historical} {Topographic} {Map} {Series} {Using} {Weakly} {Supervised} {Convolutional} {Neural} {Networks}, IEEE Access, 8, 6978--6996, \doi{10.1109/ACCESS.2019.2963213}, conference Name: IEEE Access, 2020.

\bibitem[{Uhl et~al.(2021)Uhl, Leyk, Li, Duan, Shbita, Chiang, and Knoblock}]{Uhl_Leyk_2021}
Uhl, J.~H., Leyk, S., Li, Z., Duan, W., Shbita, B., Chiang, Y.-Y., and Knoblock, C.~A.: Combining Remote-Sensing-Derived Data and Historical Maps for Long-Term Back-Casting of Urban Extents, Remote Sensing, 13, 3672, \doi{10.3390/rs13183672}, 2021.

\bibitem[{Vaswani et~al.(2017)Vaswani, Shazeer, Parmar, Uszkoreit, Jones, Gomez, Kaiser, and Polosukhin}]{NIPS2017_attn}
Vaswani, A., Shazeer, N., Parmar, N., Uszkoreit, J., Jones, L., Gomez, A.~N., Kaiser, L.~u., and Polosukhin, I.: Attention is All you Need, in: Advances in Neural Information Processing Systems, vol.~30, Curran Associates, Inc., \urlprefix\url{https://proceedings.neurips.cc/paper_files/paper/2017/file/3f5ee243547dee91fbd053c1c4a845aa-Paper.pdf}, 2017.

\bibitem[{Wu et~al.(2022{\natexlab{a}})Wu, Heitzler, and Hurni}]{wu_closer_2022}
Wu, S., Heitzler, M., and Hurni, L.: A Closer Look At Segmentation Uncertainty of Scanned Historical Maps, The International Archives of the Photogrammetry, Remote Sensing and Spatial Information Sciences, XLIII-B4-2022, 189--194, \doi{10.5194/isprs-archives-XLIII-B4-2022-189-2022}, 2022{\natexlab{a}}.

\bibitem[{Wu et~al.(2022{\natexlab{b}})Wu, Heitzler, and Hurni}]{wu_leveraging_2022}
Wu, S., Heitzler, M., and Hurni, L.: Leveraging uncertainty estimation and spatial pyramid pooling for extracting hydrological features from scanned historical topographic maps, GIScience \& Remote Sensing, 59, 200--214, \doi{10.1080/15481603.2021.2023840}, 2022{\natexlab{b}}.

\bibitem[{Wu et~al.(2023)Wu, Schindler, Heitzler, and Hurni}]{wu_domain_2023}
Wu, S., Schindler, K., Heitzler, M., and Hurni, L.: Domain adaptation in segmenting historical maps: {A} weakly supervised approach through spatial co-occurrence, ISPRS Journal of Photogrammetry and Remote Sensing, 197, 199--211, \doi{10.1016/j.isprsjprs.2023.01.021}, 2023.

\bibitem[{Yuan et~al.(2023)Yuan, Cheng, Yang, and Sester}]{Yuan_gevbev2023}
Yuan, Y., Cheng, H., Yang, M.~Y., and Sester, M.: Generating Evidential BEV Maps in Continuous Driving Space, ISPRS Journal of Photogrammetry and Remote Sensing, 204, 27--41, \doi{https://doi.org/10.1016/j.isprsjprs.2023.08.013}, 2023.

\bibitem[{Yuan et~al.(2025)Yuan, Thiemann, and Sester}]{yuan2025hismap}
Yuan, Y., Thiemann, F., and Sester, M.: Semantic Segmentation for Sequential Historical Maps by Learning from Only One Map, Preprint, 30, \urlprefix\url{https://arxiv.org/abs/2501.01845}, 2025.

\end{thebibliography}

\clearpage
\subsection*{Appendix}\label{sec:append}

\begin{minipage}{\textwidth}
    \centering
    \includegraphics[width=0.48\linewidth]{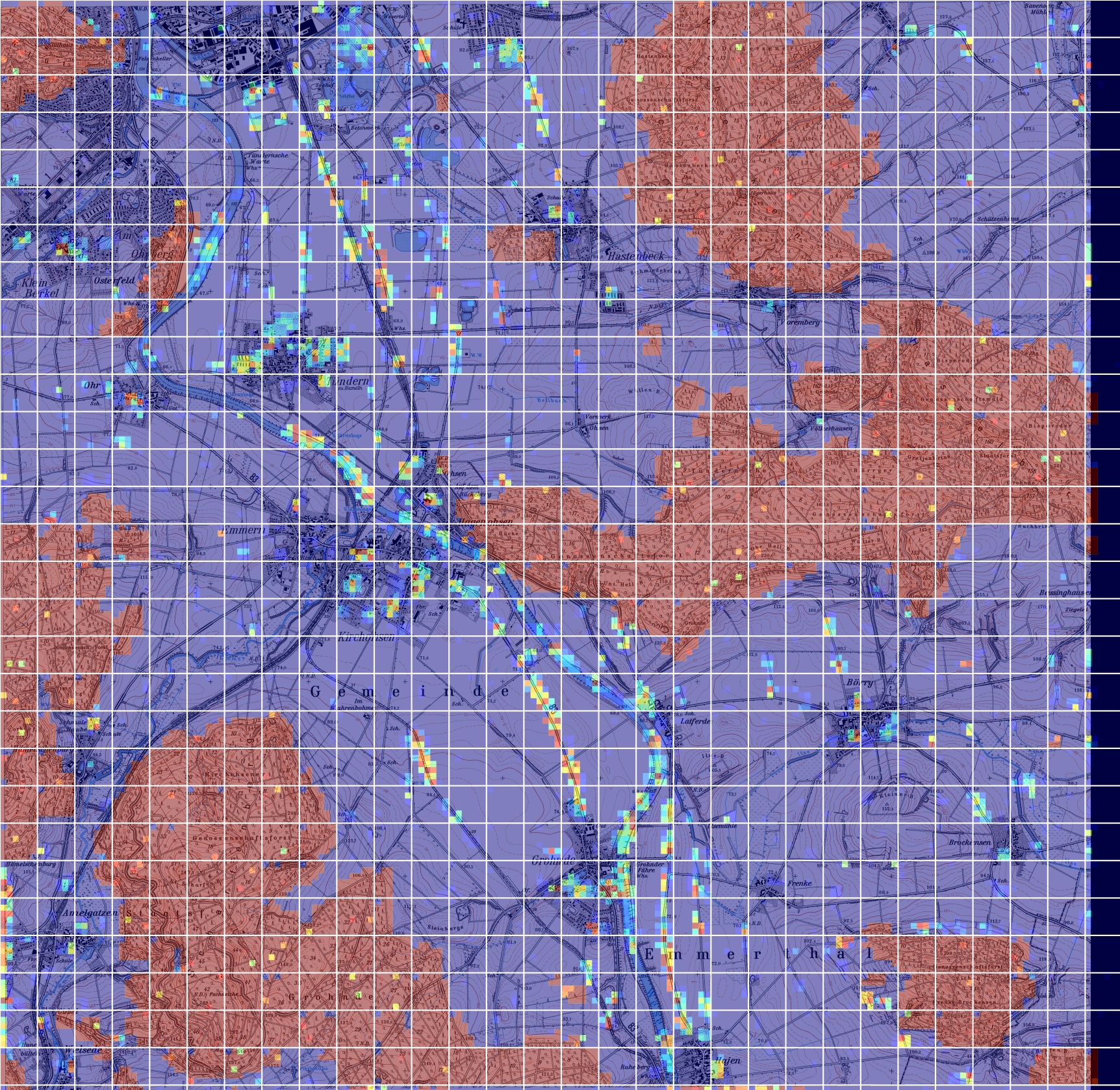}
    \hfill
    \includegraphics[width=0.48\linewidth]{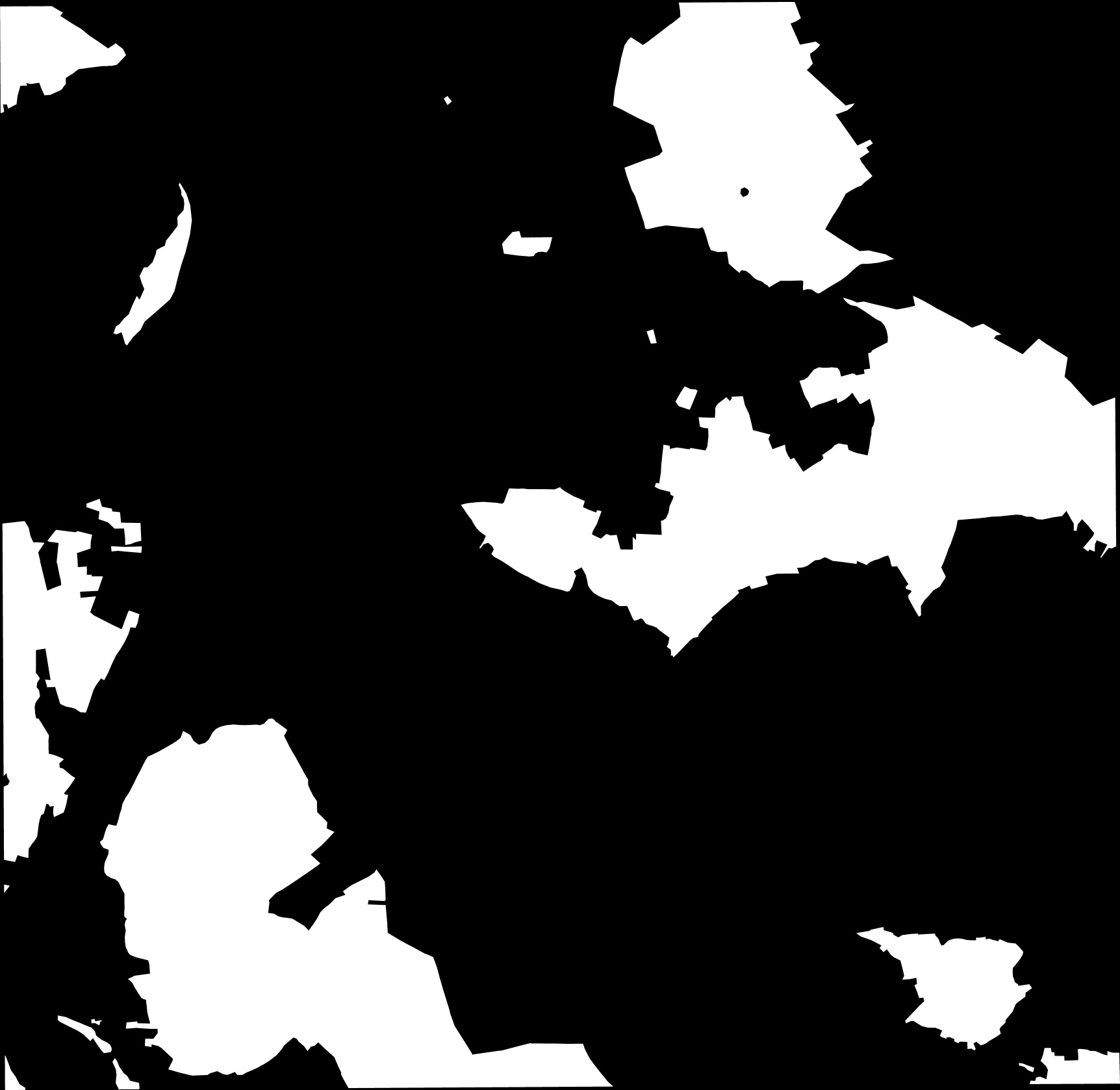}
    \captionof{figure}{Attention map for whole map sheet of \textit{Wood} class compared to ground-truth semantic labels. Left: Attention mask overlaid on historical map. Right: Ground-truth labels for \textit{Wood} class.}
\end{minipage}

\vspace{1cm}

\begin{minipage}{\textwidth}
    \centering
    \includegraphics[width=0.48\linewidth]{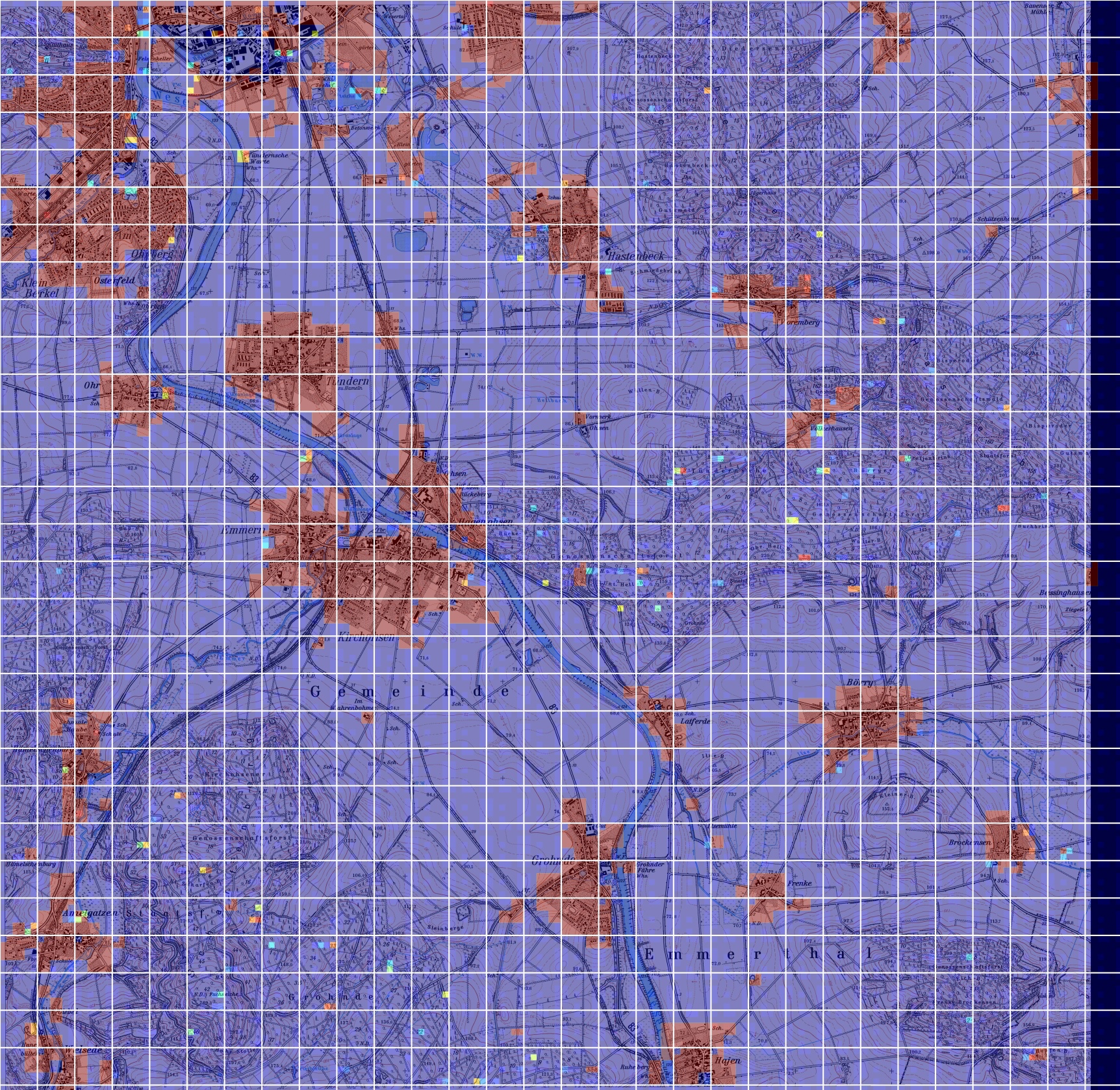}
    \hfill
    \includegraphics[width=0.48\linewidth]{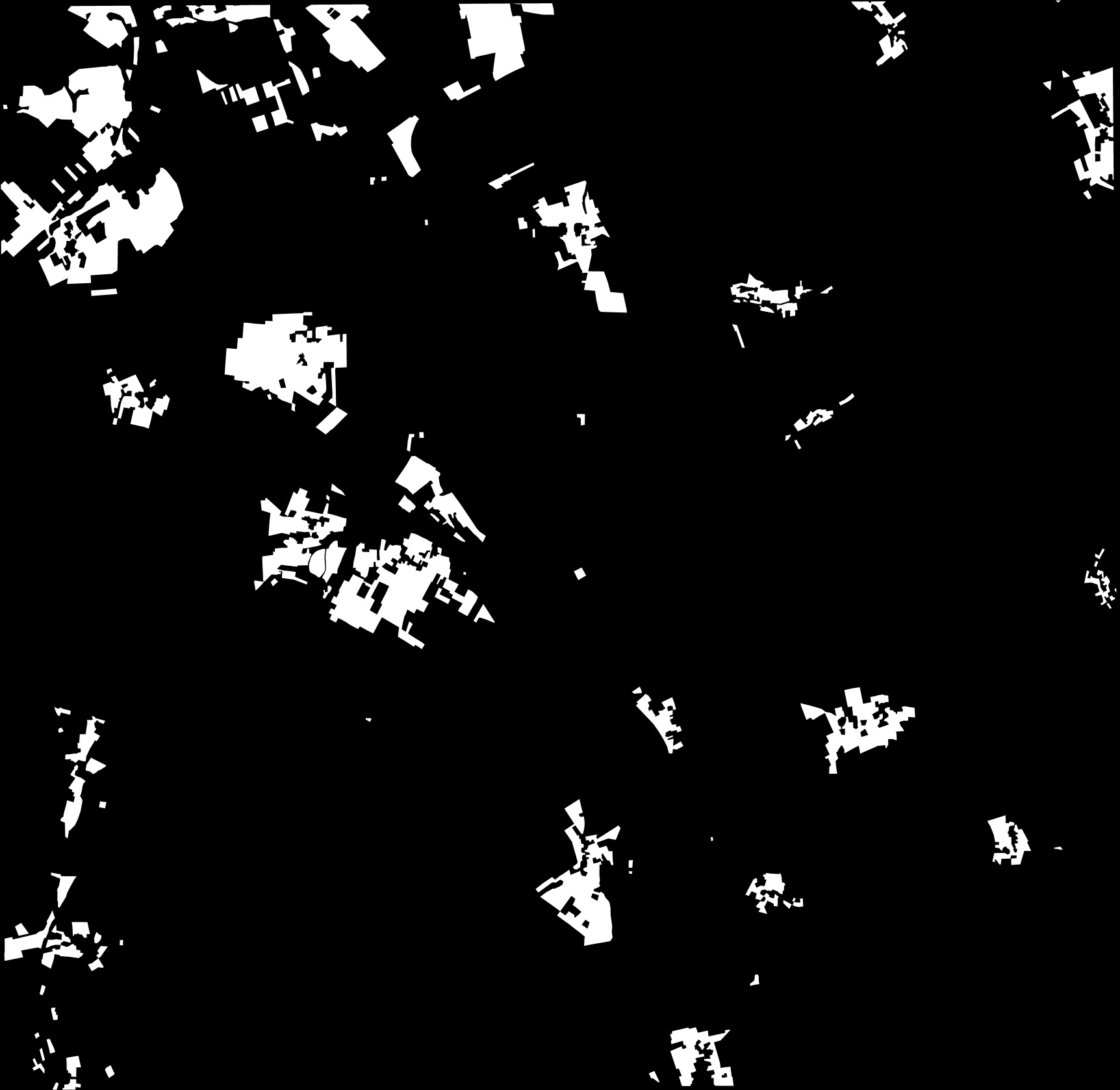}
    \captionof{figure}{Attention map for whole map sheet of \textit{Settlement} class compared to ground-truth semantic labels. Left: Attention mask overlaid on historical map. Right: Ground-truth labels for \textit{Settlement} class.}
\end{minipage}


\end{document}